\documentclass[letterpaper, 10 pt, conference]{ieeeconf}  

\IEEEoverridecommandlockouts                              

\usepackage{epsfig} 
\usepackage{mathptmx} 
\usepackage{times} 
\usepackage{amsmath} 

\usepackage{graphics} 
\usepackage{graphicx} 
\usepackage{underscore} 
\usepackage{color} 
\usepackage{booktabs} 

\usepackage{bbding}
\usepackage{pifont}
\usepackage{wasysym}

\title{\LARGE \bf
Answerability Fields: Answerable Location Estimation \\
via Diffusion Models
}

\author{Daichi Azuma$^{1}$, Taiki Miyanishi$^{2}$, Shuhei Kurita$^{3}$, Koya Sakamoto$^{2,4}$ and Motoaki Kawanabe$^{2}$
\thanks{$^{1}$ Sony Semiconductor Solutions}%
\thanks{$^{2}$ Advanced Telecommunications Research Institute International (ATR), Kyoto, Japan}%
\thanks{$^{3}$ RIKEN Center for Advanced Intelligence Project, Tokyo, Japan}%
\thanks{$^{4}$ Graduate School of Informatics, Kyoto University, Japan}%
}

\begin{document}

\maketitle
\thispagestyle{empty}
\pagestyle{empty}

\begin{abstract}
In an era characterized by advancements in artificial intelligence and robotics, enabling machines to interact with and understand their environment is a critical research endeavor. 
In this paper, we propose Answerability Fields, a novel approach to predicting answerability within complex indoor environments. 
Leveraging a 3D question answering dataset, 
we construct a comprehensive Answerability Fields dataset, encompassing diverse scenes and questions from ScanNet. 
Using a diffusion model, we successfully infer and evaluate these Answerability Fields, demonstrating the importance of objects and their locations in answering questions within a scene. 
Our results showcase the efficacy of Answerability Fields in guiding scene-understanding tasks, laying the foundation for their application in enhancing interactions between intelligent agents and their environments.
\end{abstract}
\section{Introduction}
\label{sec:intro}
The rapid advancements in applying deep neural networks to embodied agents have enabled capabilities such as navigating indoor environments following linguistic instructions~\cite{vlnAnderson2018CVPR,batra2020objectnav, objgoalchaplot2020object}, 
dexterous object manipulation~\cite{zeng2020transporter,akkaya2019solving,kalashnikov2018scalable}, 
and answering questions within 3D environments~\cite{eqaMatterport,ma2022sqa3d}.
In particular, the capacity of embodied agents to interpret and respond to queries within 3D environments is essential for developing robots that can comprehend human language and execute tasks accordingly.
In existing methods, agents randomly placed in an unknown environment must explore it to find specific objects for providing accurate answers to posed questions~\cite{embodiedqa}.
However, even though the layouts of indoor environments are often known in real-world scenarios and robots typically possess their own maps, existing agent-perspective question answering (QA) methods do not utilize them.
The question naturally arises, ``Can we use an indoor 2D map to answer the question about 3D space?''
There are the models to predict the location of objects in 3D space from 2D map which agent generated~\cite{ramakrishnan2022poni, Gadre2022CLIPOW}
In the QA task, the agent need to find a location where agent can visually acquire not only the target object location in 3D space, but also the necessary information to answer the question, i.e., these are not only the target objects, but also other objects and object-position relationships included in the question.
We focused on question answering using layouts and further investigated what locations the agent would need to spawn for QA.

\begin{figure}[t]
\begin{center}
\includegraphics[keepaspectratio, scale=0.43]{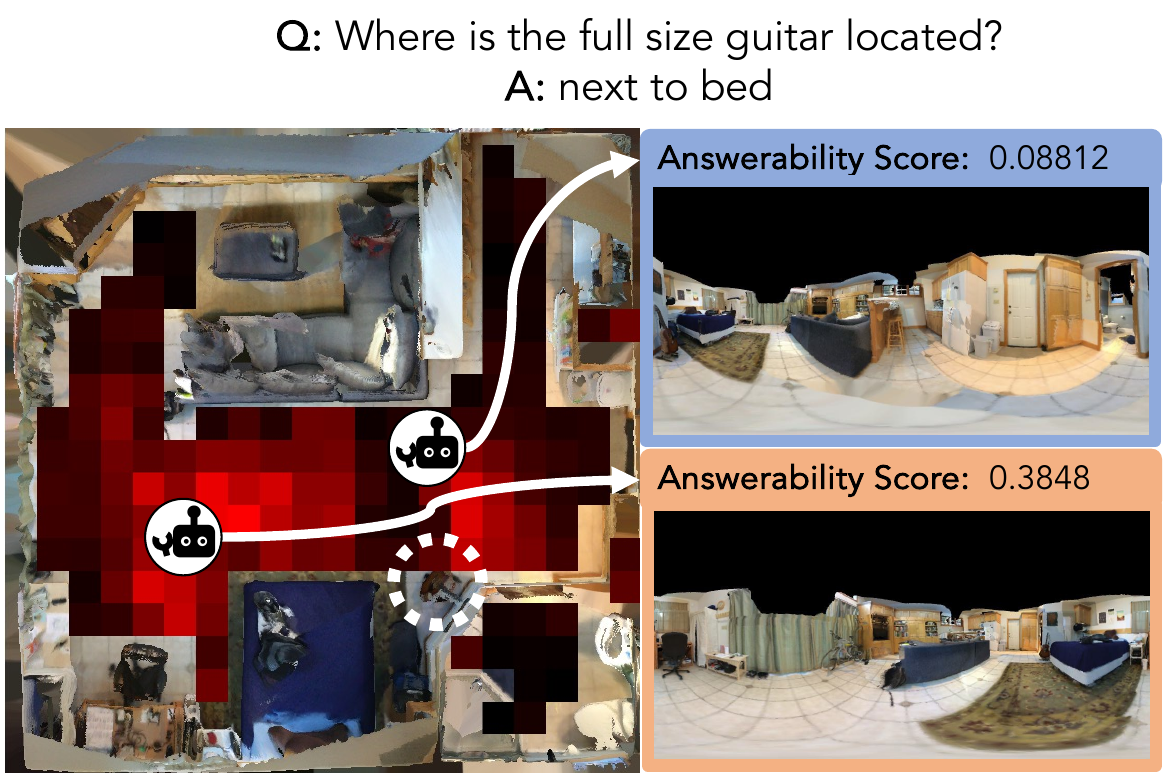}
\end{center}
\vspace{-0.4cm}
\caption{
We propose Answerability Fields to make agent efficiently understanding scenes. We compute the score of answerability to the questions in each location by using strong Visual Question Answering (VQA) model called OFA.}
\label{fig:teaser}
\vspace{-0.3cm}
\end{figure}
In this work, we propose Answerability Fields (AnsFields) that facilitate robots answering questions in 3D environments using indoor 2D maps.
AnsFields is a grid representation where each cell indicates the possibility that the robot will answer a given question at that location.
AnsFields is created by the vision and language foundation model called OFA~\cite{wang2022ofa}.
We calculated the probability that OFA would return an appropriate answer based on the question and the agent-perspective image obtained from that position when the agent was spawned in each grid.
Figure~\ref{fig:teaser} illustrates an example of AnsFields for the question ``Where is the full size guitar located?'', where locations with high answerability have higher scores, and locations with low answerability have lower scores.
In this example, in addition to ``guitar'' which is the subject of the answer, an image containing context that shows where the guitar is located is required. The score is higher for locations that can be seen in the relationship between the ``bed'' and ``guitar''.
In contrast to previous embodied QA works~\cite{eqaMatterport,ma2022sqa3d}, once the location of the high answerability is known based on AnsFields, the question can be answered with high accuracy by moving to that location and conducting visual question answering (VQA) based on the surrounding images there.
Additionally, we present a diffusion model tailored to predict AnsFields based on top-down view images of the environment and questions.
It employs a technique known as InstructPix2Pix~\cite{brooks2022instructpix2pix} designed for image-to-image translation tasks, guided by textual instructions.
We train a diffusion model with paired data that includes questions, top-down view image inputs, and images of AnsFields outputs.
Agent-perspective QA is performed on the highest-scoring location of the  AnsFields predicted by the text-conditioned diffusion model.

We demonstrate that AnsFields created using the ScanQA dataset~\cite{Azumascanqa}, a 3D Question Answering dataset using scenes from richly-annotated 3D scans of ScanNet~\cite{scannetdata}, are effective for agent-perspective QA.
Specifically, our AnsFields significantly improve the QA performance by more than 5 $\%$ compared to answering in front of a target object, which is a primary objective of previous works~\cite{embodiedqa,eqaMatterport}.

We also demonstrate that our text-conditioned diffusion model predicts the AnsFields that enhance agent-perspective QA performance, which outperforms the 3D QA method~\cite{Azumascanqa} and the methods conducting QA at random locations.

\section{Related Work}
~\begin{figure*}
\begin{center}
\includegraphics[keepaspectratio, scale=0.5]{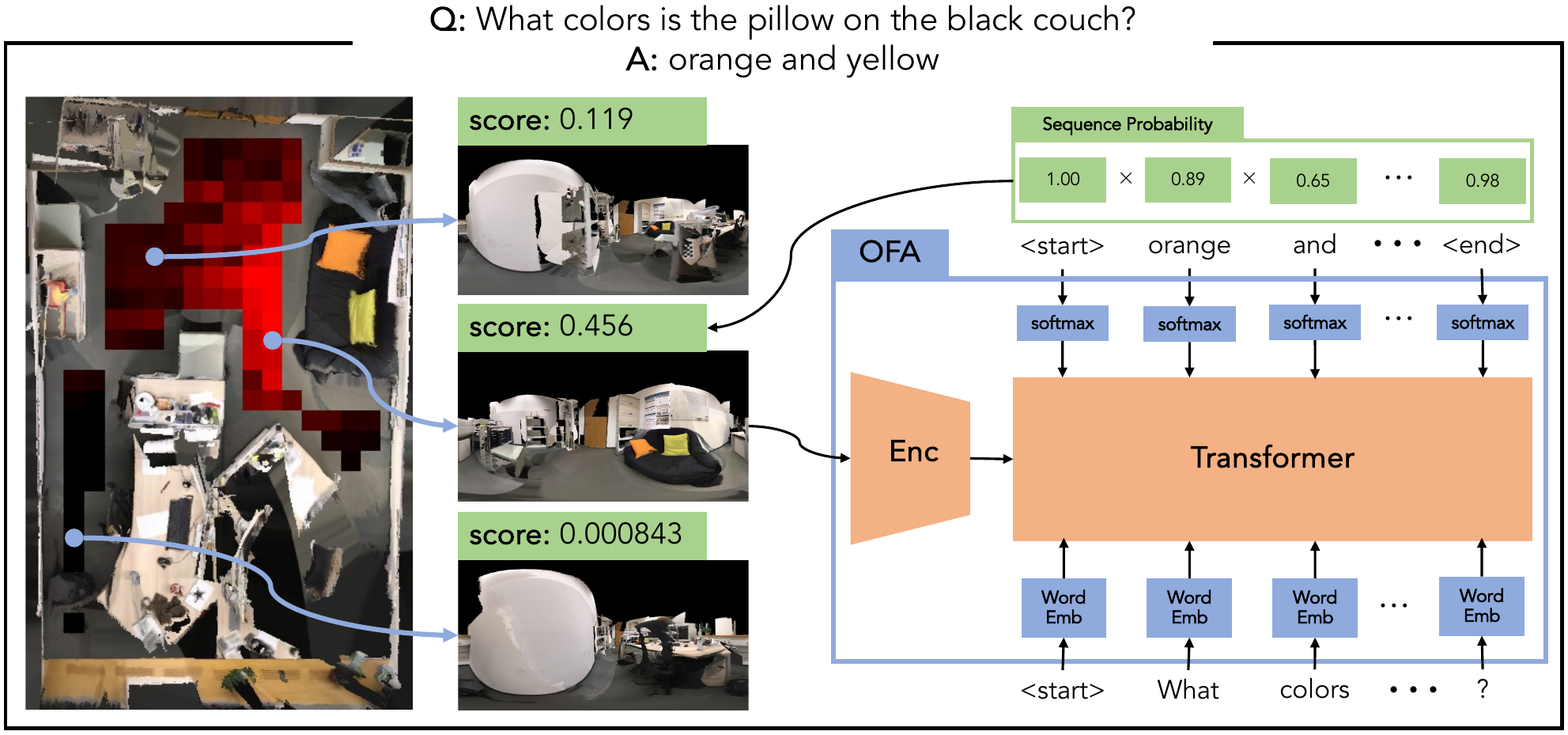}
\end{center}
\vspace{-0.4cm}
\caption{
Represents a method for generating answerability fields, which calculates the probability for each sequence to return an appropriate answer when a question and a image are entered into VQA model.}
\label{fig:answerability}
\vspace{-0.3cm}
\end{figure*}
\label{sec:relatedwork}
\noindent \textbf{Question Answering in 3D Space.}
EQA serves as the convergence point of visual navigation and visual question answering (VQA) within intricate 3D environments~\cite{embodiedqa, eqaMatterport}. 
In EQA, agents are tasked with navigating their surroundings to identify objects and provide accurate responses to posed questions. This task necessitates not only perceptual capabilities but also spatial reasoning and contextual understanding, underscoring the intricacy of embodied interactions in rich environments.

ScanQA~\cite{Azumascanqa} utilizes 3D point cloud data representing indoor environments. Question and answer pairs in ScanQA are derived from the ScanRefer dataset~\cite{chen2020scanrefer}, which contains human-annotated descriptions of objects from ScanNet scenes~\cite{scannetdata}. These descriptions are automatically converted into questions and further refined using Amazon Mechanical Turk (AMT). The corresponding answers are also manually annotated via AMT. As ScanQA is manually created from 3D scans, this QA dataset is well-suited for real-world applications.

\noindent \textbf{Generative Models for Map Generation.}
In recent years, diffusion models have significantly improved in accuracy and can generate appropriate images under various conditions. Combined with advancements in Large Language Models (LLMs), it is now possible to include text, audio, and images as conditions and generate outputs. Noteworthy examples include InstructPix2Pix, Codi, and Controlnet~\cite{brooks2022instructpix2pix, zheng2023minigpt5, zhang2023adding, tang2023anytoany, Emu2}. These systems not only generate conditioned data but also comprehend the data and solve various tasks. For instance, DiffusionInst~\cite{DiffusionInst} and DiffusionDet~\cite{chen2022diffusiondet} utilize diffusion models to recognize objects in images, predict segmentation, and bounding boxes, outperforming existing methods

Diffusion models have also found application in scene understanding and robotics. MapPrior~\cite{zhu2023mapprior} employs a map image as input and utilizes a diffusion model to classify classes within the map. Models predicting the trajectory of an agent, such as PolyGRAD~\cite{rigter2023world} and DiPPeR~\cite{liu2024dipper}, leverage classifier guidance based on state and action. Look Outside the Room~\cite{ren2022look} generates an image of the agent's path from an image of an indoor environment. Additionally, there are models that generate house layouts using GANs, enabling scene understanding with diffusion models~\cite{nauata2021house}.

\noindent \textbf{Semantic Fields.}
Semantic perception in the realm of embodied artificial intelligence involves understanding and interpreting the environment in terms of meaningful concepts and relations. Several recent advancements contribute to this area.

CLIP-Fields~\cite{shafiullah2022clipfields} introduces an implicit scene model capable of tasks such as segmentation, instance identification, semantic search over space, and view localization. By learning a mapping from spatial locations to semantic embedding vectors, CLIP-Fields~\cite{shafiullah2022clipfields} can perform these tasks without direct human supervision, outperforming traditional methods like Mask-RCNN~\cite{he2018mask} in tasks such as a few-shot instance identification or semantic segmentation with only a fraction of the examples. Additionally, when used as a scene memory, CLIP-Fields~\cite{shafiullah2022clipfields} enables robots to navigate semantically in real-world environments, demonstrating practical utility.

PONI~\cite{ramakrishnan2022poni} addresses the challenge of ObjectGoal navigation by separating the skills of 'where to look?' and 'how to navigate to (x, y)?'. By treating 'where to look?' as a perception problem, PONI~\cite{ramakrishnan2022poni} learns to predict potential functions from semantic maps, facilitating efficient navigation to unseen objects. This modular approach achieves state-of-the-art performance on ObjectNav tasks while significantly reducing computational costs for training, making it promising for real-world deployment.

VLMAP~\cite{huang23vlmaps} proposes a spatial map representation integrating pre-trained visual-language features with 3D reconstructions of physical environments. This approach allows natural language indexing of maps without additional labeled data, enabling more precise navigation according to complex language instructions. VLMAP~\cite{huang23vlmaps} facilitates autonomous map generation from video feeds, enhancing robot navigation capabilities in both simulated and real-world environments.

ConceptFusion~\cite{conceptfusion} introduces a scene representation that is open-set and multimodal, enabling reasoning about concepts across various modalities such as natural language, images, and audio. Leveraging foundation models pre-trained on internet-scale data, ConceptFusion~\cite{conceptfusion} achieves effective zero-shot spatial reasoning and retains long-tailed concepts better than supervised approaches. Extensive evaluations demonstrate its efficacy across different real-world applications, offering new possibilities for blending foundation models with 3D multimodal mapping.

These advancements collectively contribute to enhance semantic perception in embodied artificial intelligence systems, enabling more robust and contextually aware interactions with complex environments.

\section{Problem Fomulation}
In this work, we address the task of agent-perspective QA in the 3D space, where a robot is placed within a previously seen 3D environment, moves to the best location for QA, and provides accurate responses to textual questions using the robot's egocentric RGB images.
In contrast to the existing embodied QA tasks~\cite{embodiedqa, eqaMatterport, eqamodular}, the robot already has the 2D map (top-down image) of the placed environment that can be used for this task.
Thus, using a map to find a favorable location for answers is important for solving this task.

\if[]
In the realm of Navigation and Visual Question Answering (VQA), achieving both semantic and spatial understanding of the scene is paramount. It is essential to discern not only where information is universally present within the scene but also to predict where it can be accessed. In VQA, answering a question requires the ability to identify what the question is asking and also where the relevant information is available to answer it. We hypothesize that we can predict the most efficient locations for retrieving this information based on the objects in the scene and their spatial arrangement. For instance, being too close or too far from the object in question may not provide the best viewpoint; instead, having a bird's-eye view of the entire object is crucial for understanding the context of the question. To explore this concept, we developed the Answerability Fields dataset to assess whether we can effectively capture the distribution of answerability at different spatial locations within the environment.

By integrating Answerability Fields into the EQA framework, we anticipate several benefits. Firstly, it enhances the agent's ability to navigate and comprehend complex environments by providing valuable insights into the spatial distribution of answerability. Secondly, it facilitates more informed decision-making processes, enabling agents to prioritize actions and locations based on their likelihood of yielding relevant information. Ultimately, Answerability Diffusion contributes to the advancement of embodied artificial intelligence by fostering more efficient and contextually aware interactions between intelligent agents and their environments.
\fi

\section{Approach}
\label{sec:method}
In this section, we introduce the Answerability Fields (AnsFields) that represent the most efficient locations for answering questions in the 3D scenes.
We hypothesized being too close or too far from the object related to the answer may not provide the best viewpoint; 
instead, it is crucial to view the object and its surrounding context to identify what the question is asking and determine where the relevant information is available to answer it.
To explore the concept, we propose the AnsFields.
First, we introduce the definition of AnsFields, then describe how to create AnsFields using vision-and-language foundation models, and then present the diffusion models for predicting AnsFields in unseen 3D environments.

\subsection{Definition of Answerability Fields}
Answerability Fields (AnsFields) is the grid representation of answerability for embodied agents.
Answerability is calculated as the ability to provide appropriate answers to questions about panoramic images and scenes acquired at each location. 
As depicted in Figure~\ref{fig:teaser}, each cell within the grid is assigned answerability scores, 
where the higher probability of providing the correct answer to a question, while a value closer to 0 suggests a lower probability.
We consider the probability of a visual question answering (VQA) at each cell as the answerability score.

\subsection{Creation of Answerability Fields}
To compute the AnsFields, we employed the powerful pre-trained transformer-based encoder-decoder framework, OFA~\cite{wang2022ofa}, which was trained on a large corpus of image-text pair data, serving as the VQA model.
It achieves remarkable performance in a series of cross-modal tasks such as visual grounding, VQA, and image captioning.
Here, while any model can be used as the VQA model, in this work, we utilize OFA due to its high performance while being operable on a single GPU.
Initially, we fine-tuned the OFA pre-trained model with the 3D-QA dataset, which takes a question and a panoramic image at each location on the grid of indoor scenes as input and predicts the token sequence of the answer.
Then, OFA computes the answerability score, which represents how well a VQA model predicts answer tokens. This score is obtained as the probability of generating the correct answer tokens.
We calculate scores for all grid cells within the range an agent can move (i.e., NavMesh area) to generate the AnsFields.
Figure~\ref{fig:answerability} depicts an example of how to compute the AnsFields within the range an agent can move. 
When addressing the question ``What color is the pillow on the black couch?'' with the answer ``orange and yellow,'' the highest score is assigned to the location that captured a panoramic image showcasing both the orange and yellow pillows. 
Conversely, lower scores are given to locations that obtained images not clearly showing both pillows or not showing them at all.
Using the VQA model, AnsFields can quantitatively express locations that are easier to answer.

\begin{figure}
\begin{center}
\includegraphics[keepaspectratio, scale=0.35]{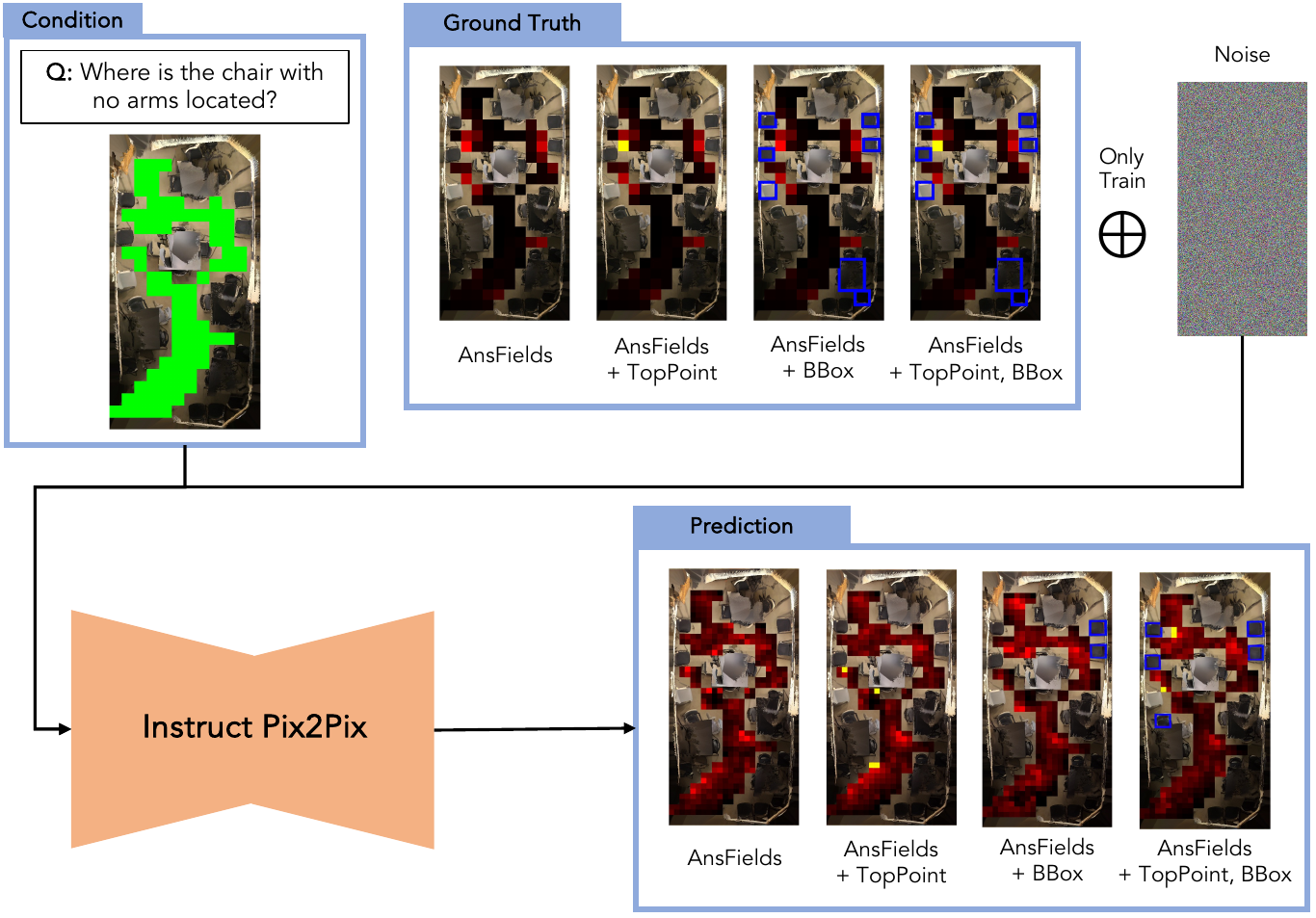}
\end{center}
\vspace{-0.4cm}
\caption{
This is the overview of generating Answerability Fields via Instruct Pix2Pix. On train, learning to predict noise by adding noise to the correct image. The correct images were trained and comparatively evaluated with only Answerability drawn and, TopPoints highlighted and BoundingBoxes on the top-down view images of the scene}
\label{fig:diffusionoverview}
\vspace{-0.3cm}
\end{figure}
\subsection{Generation of Answerability Fields by Diffusion Models}
We generate the AnsFields for unknown questions
using a diffusion model called InstructPix2Pix~\cite{brooks2022instructpix2pix}.
It is an extension of the latent diffusion model (LDM)~\cite{rombach2021highresolution}, which samples data in latent space via a variational autoencoder~\cite{kingma2022autoencoding} with UNet~\cite{ronneberger2015unet} architecture.
Formally, given an RGB image $C_{i}$ and text prompt $C_{p}$, the InstructPix2Pix model processes a noisy image $Z_{t}$ as its input and produces a new image $Z_{0}$ using the Classifier-Free Guidance~\cite{ho2022classifierfree}. 
In the Classifier-Free Guidance, 
the diffusion model predicts the amount of noise present in the input image $Z_{t}$, using the denoising U-Net $\tilde{e}{\theta}$ which is divided into two networks, conditional noise predictions $e_{\theta}(Z_{t}, c)$ and unconditional noise predictions $e_{\theta}(Z_{t}, \phi, \phi)$ as:
\begin{equation*}
\begin{split}
\tilde{e}{\theta}(Z{t}, C_{i}, C_{p}) &= e_{\theta}(Z_{t}, \phi, \phi) \\
&+ S_{i} \cdot (e_{\theta}(Z_{t}, C_{i}, \phi) - e_{\theta}(Z_{t}, \phi, \phi)) \\
&+ S_{t} \cdot (e_{\theta}(Z_{t}, C_{i}, C_{p}) - e_{\theta}(Z_{t}, C_{i}, \phi))
\end{split}
\end{equation*}
where $S_{i}$ and $S_{t}$ are the scale factors in adding up the conditional scores of the image and text, respectively. 
A higher scale factors indicate a stronger influence of the condition on the generated image.
In our experiments, the image condition scale factor $S_{i}$ is set to 1.5 and the image and text condition scale factor is set to 7.

Using InstructPix2Pix, our method learns to predict the AnsFields conditioned on textual questions and top-down view images of indoor scenes.

\noindent \textbf{Highlight Top Answerability Score Point.}
In AnsFields, answerability is normalized and plotted based on the size of the R value of the RGB value (0 to 255). Therefore, the RGB value of the point with the highest answerability is represented as $[255, 0, 0]$. By explicitly learning the points where the most visual information to answer the question is gathered, we anticipated an increase in inference accuracy.

\noindent \textbf{Visualize Object BBox.}
To infer answerability, understanding the location of the target objects mentioned in the question and the relationship between their locations is considered necessary. Hence, bounding boxes were applied with RGB values $[0, 0, 255]$ to the objects related to the question in the training image. By allowing the model to infer the location of the target object along with the answer possibilities, we aimed to enhance the understanding of the environment and measure the improvement in accuracy.

The overview of AnsFields predictions in InstructPixPix is shown in Fig~\ref{fig:diffusionoverview}.
Random noise is added to the correct AnsFields image and, as a further condition, the question and scene top-down view image are added to predict the noise with UNet~\cite{ronneberger2015unet}, and compute the Mean Square Error (MSE) with the prediction results and correct image to train the model. The correct AnsFields image is trained with each of the above conditions: only AnsFields are added to the top-down view of the scene, TopPoints are illustrated and the BoundingBox is annotated. During sampling, noise images, questions, and scene images are used to generate AnsFields images.

\section{Experiments}
\label{sec:experiments}
\subsection{Datasets}
\subsubsection{Images of AnsFields}
As a 3D-QA dataset, we utilized the indoor scenes from the ScanNet dataset~\cite{scannetdata} and obtained question-answer data related to these scenes from the ScanQA dataset~\cite{Azumascanqa}.
The AnsFields' images are created according to the following steps.
(i) We filled in missing parts of the floor in the point cloud of ScanNet data and then meshed the point cloud using Delaunay triangulation.
(ii) We identified navigable areas corresponding to the floor and divided them into a grid based on the scene size. 
(iii) We then employed the VQA model to compute answerability using panoramic images at the grid locations.
(iv) We plotted the normalized answerability score on this top-down view image created from the scene's point cloud.
The image of AnsFields was represented by the R color value of RGB, which the diffusion model tries to predict.
We excluded scenes with navigable areas that were too small. Consequently, we created AnsFields images for 619 scenes and 31,741 question-answer pairs.

\subsubsection{Estimation of the Best Viewpoint}
From the image inferred by the model, we calculated the point with the highest score of answerability to the question and obtained a panoramic image of the environment taken at that location. Specifically, we estimated the location of navigable areas in the generated images using a transformation matrix between the pre-prepared scene and the top-down view images. The point with the highest RGB value of R in that area was then identified. Furthermore, we calculated the position of that point in the scene using the transformation matrix, placed the agent, and acquired the panoramic image. In essence, this point represents the most probable position to answer the question inferred by the model, allowing us to validate whether the correct inference was made by evaluating the image of that point.

\subsection{Implementation Details}
We utilized a newly created collection of AnsFields images and question and answer pairs on the ScanQA~\cite{Azumascanqa} dataset to train the InstructPix2Pix model~\cite{brooks2022instructpix2pix}.
The model was trained using AdamW~\cite{loshchilov2019decoupled} with an initial learning rate of 1e-5 and a batch size of 16 for 150 epochs with train data.
Subsequently, AnsFields were generated on the test data.

\subsection{Evaluation}
To verify the accuracy of the locations predicted by AnsField as the most likely to contain the answer to a question, a Visual Question Answering (VQA) test was conducted using a panoramic image taken at the predicted location.
We used EM@1 and EM@10 for the VQA metric, where EM@$K$ represents the percentage of predictions where the top $K$ predicted answers exactly match one of the ground-truth answers.

\subsection{Baselines}
We compared the AnsFields inferred by our diffusion model against various other methods. 
Note that no existing models are specifically tailored for the inference of AnsFields.
Therefore, we selected a range of closely related methods for this comparison.

\noindent \textbf{Top-Down View VQA.}
We utilized mesh images captured from a top-down view (referred to as TopDownQA) to capture the entire room with a single image for performing 2D-QA.
Given a question and top-down view image, the method directly produces the answer.
We use the same VQA model, OFA, as the proposed method to see whether an agent-perspective panoramic or top-down view image is more effective.

\noindent \textbf{3D-QA.}
For comparison, we introduce a 3D-QA method that employs a point cloud of the entire scene to answer questions. 
Unlike 2D-QA which uses top-down images, 3D-QA can consider 3D spatial relationships of objects in a room.
We use ScanQA~\cite{Azumascanqa}, the most commonly used method for 3D-QA on indoor scenes.

\noindent \textbf{VQA at Random Location.}
To investigate whether it is effective to VQA at appropriate positions in agent-perspective QA, we will prepare a method to QA at random positions
(referred to as Random Spawn.)

\noindent \textbf{VQA on Top-down View Attention.}
This method uses the text-image matching model BLIP~\cite{li-etal-2023-lavis} to determine the most salient location within the Navmesh area in a top-down view image based on a given question using the attention mechanism.
Subsequently, an agent is positioned at this identified location to capture a panoramic image.
The image is used for conducting agent-perspective QA using the OFA model~\cite{wang2022ofa}.
\section{Results}
\label{sec:results}
\begin{table}
\begin{center}
    \caption{
        VQA Performance comparison to the existing method.
        }
	\normalsize\begin{tabular}{lcc}
        \toprule
MODEL & EM@1 & EM@10 \\
\midrule
\midrule
\multicolumn{1}{l}{\textbf{2D-QA and 3D-QA}} \\
TopDownQA & 35.00 & 71.32 \\
ScanQA          & 31.18 & 68.35 \\
\midrule
\multicolumn{1}{l}{\textbf{Agent-perspective QA}} \\
Random Spawn  & 37.75 & 73.17 \\
Top-down View Attention  & 38.38   & 74.36 \\
AnsFields (Ours)  & \textbf{38.77} & \textbf{74.41}   \\
\bottomrule
	\end{tabular}

    \label{table:baseline}
\end{center}
\vspace{-4mm}
\end{table}

\if[]
\begin{table}
\begin{center}
	\normalsize\begin{tabular}{lcc}
        \toprule
MODEL & EM@1 & EM@10 \\
\midrule
\midrule
\multicolumn{1}{l}{\textbf{2D-QA and 3D-QA}} \\
ScanQA           & 31.18 & 68.35 \\
OFA (top-down view image) & 35.00 & 71.32 \\
\midrule
\multicolumn{1}{l}{\textbf{Agent-perspective QA}} \\
Random Position    & 37.75 & 73.17 \\
Blip               & 38.38   & 74.36 \\
Ours  & \textbf{38.77} & \textbf{74.41}   \\
\bottomrule
	\end{tabular}
    \caption{
        VQA Performance comparison to existing method.
        }
    \label{table:experiments}
\end{center}
\vspace{-4mm}
\end{table}
\fi
\begin{figure*}
\begin{center}
\includegraphics[keepaspectratio, scale=0.70]{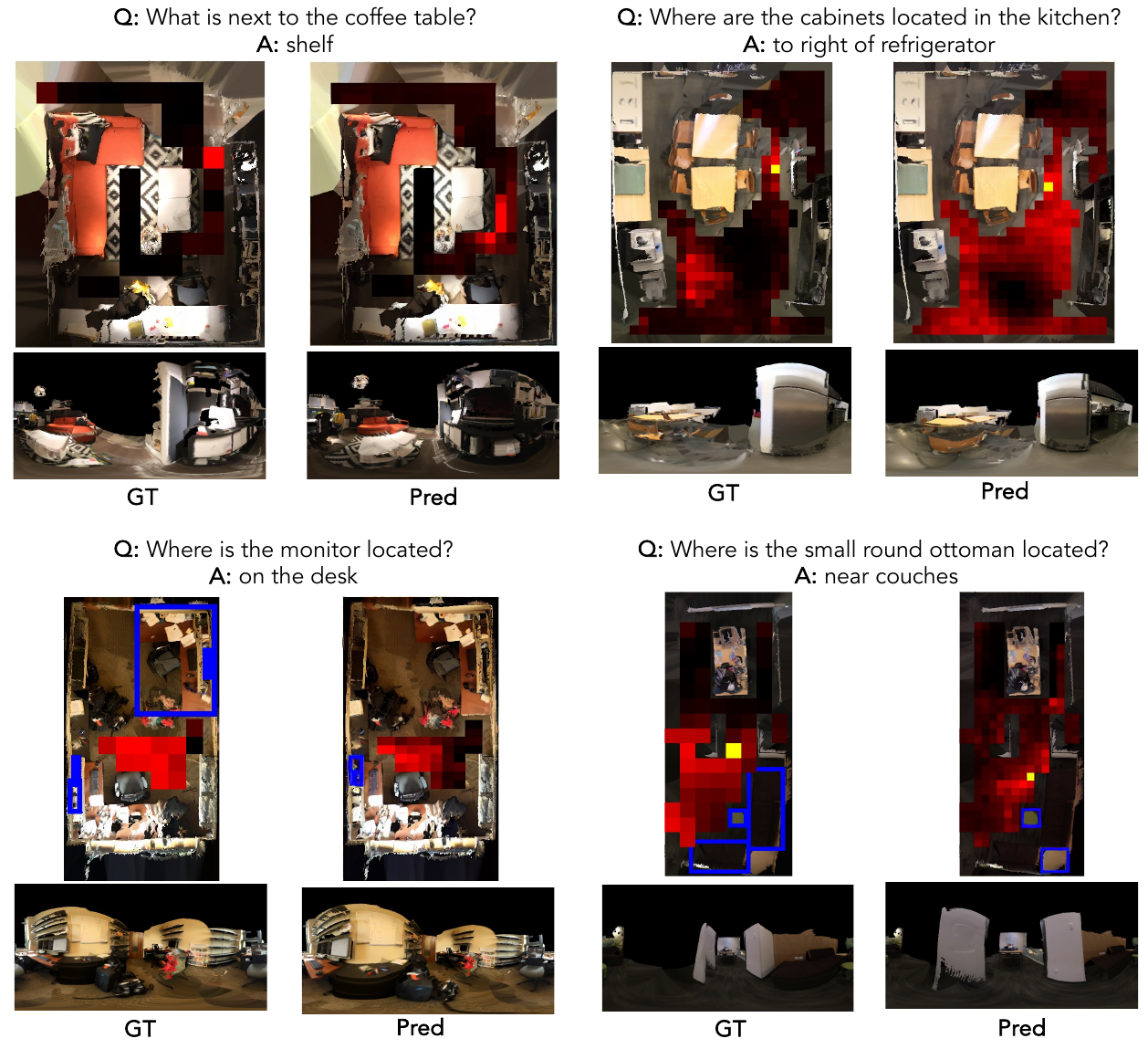}
\end{center}
\vspace{-0.4cm}
\caption{
We used InstructPix2Pix to predict answerability fields for a variety of unknown scenes. The figure shows  AnsFields to the question and the panoramic image taken in the location of the highest score of answerability.}
\label{fig:qu}
\vspace{-0.3cm}
\end{figure*}

\subsection{Comparison to Baselins}
\begin{table}
\begin{center}
\caption{
        Ablation Study of the images which input to Instrunc-Pix2Pix as conditions.
}
\normalsize\begin{tabular}{cccc}
        \toprule

TopPoint & BBox & EM@1 & EM@10 \\
\midrule
\Checkmark & & 37.85 & 73.78 \\
 & \Checkmark &  36.73 & 73.51 \\
\Checkmark & \Checkmark & \textbf{38.77} & \textbf{74.41} \\
\bottomrule
	\end{tabular}

    \label{table:ablation}
\end{center}
\end{table}

\if[]
\begin{table*}
\begin{center}
	\normalsize\begin{tabular}{cccccccccc}
        \toprule
MASK_FLOOR & HIGHLIGHT & BBOX & FID$\downarrow$ & SSIM$\uparrow$ & PSNR$\uparrow$ & MSE$\downarrow$ & PM$\downarrow$ & EM@1 & EM@10 \\
\midrule
\Checkmark &  &                      & 43.12 & \textbf{0.7235} & \textbf{21.36} & 0.2391  & 204.30 & 38.12 & 73.49 \\
\Checkmark & \Checkmark &            & 45.04 & 0.7208 & 19.91 & \textbf{0.2292} & \textbf{164.50t} & 37.85 & 73.78 \\
\Checkmark &  & \Checkmark           & \textbf{33.53} & 0.6836 & 18.94 & 0.2553 & 200.78 & 36.73 & 73.51 \\
\Checkmark & \Checkmark & \Checkmark & 39.30 & 0.6571 & 17.29 & 0.2386 & 222.57 & \textbf{38.77} & \textbf{74.41} \\
\bottomrule
	\end{tabular}
    \caption{
        Ablation Study of the images which input to Instrunc-Pix2Pix as conditions.
    }
    \label{table:ablation}
\end{center}
\end{table*}
\fi
Table~\ref{table:baseline} shows the performance of the baselines and our method on our QA dataset.
Our agent-perspective QA method significantly outperformed both the 2D-QA method using top-down view images and the 3D-QA method.
The results suggest that panoramic images from the agent's viewpoint are more effective for QA of indoor spatial reasoning than point clouds or top-down view images of the entire scene.
Additionally, our method achieving EM@1 and EM@10 38.77 $\%$ and 74.41 $\%$, respectively
outperformied ``'Random Spawn'' and ``Top-down View Attention'' with 38.38 $\%$ for EM@1 and 74.36 $\%$ for EM@10. 
This indicates that in agent-perspective QA, conducting VQA at locations relevant to the question is more effective than performing VQA at random locations.
In addition, the proposed method for generating AnsFields can predict locations more useful for QA than those predicted by image-text matching of a question and a top-down view image.

\subsection{Ablation Study}
We investigated the extent to which predicting object bounding boxes (refer to BBox) and the locations with the highest answerability scores (refer to TopPoint) affect the performance of the final agent perspective QA.
Table~\ref{table:ablation} shows the ablation results.

The results show the VQA performance with both TopPoint and BBox had the highest accuracy, 38.77 $\%$ for EM@1 and 74.41 $\%$ for EM@10. 
This result can be attributed to the diffusion model's ability to learn and predict the locations of objects and areas with high answerability scores, enabling more accurate predictions of Answerability Fields.

\subsection{Qualitative Analysis}
Finally, we show the results of the predicted answerability for various scenes and questions in Figure~\ref{fig:qu}. 
In the first example, the question ``What is next to the coffee table?'' is given and the AnsFields show the score of answerability to predict ``shelf'' as the answer. This question requires the identification of a location where the ``coffee table'' and surrounding objects are well visible. The correct panoramic image shows that an image has been obtained that clearly shows the positional relationship between ``coffee table'' and ``shelf''. However, the prediction shows that object recognition is working well, with higher scores for locations near the ``coffee table'' where the positional relationship of the surrounding objects is easy to identify. However, the panoramic image in the predicted location is difficult to recognize the ``shelf''.

In the next example, the model is required to predict the location of `kitchen cabinets''. In this example, TopPoint is highlighted in AnsFields. As with the correct answer image, the prediction results show higher scores for locations where `kitchen cabinets'' and the appropriate answer object, `refrigerator'', are properly shown, and lower scores for locations where either is not shown.

The following example is the result when train images are annotated with bounding boxes. The question asks for the location of the ``monitor''. There are several monitors in this scene and all of them are on the desk. The correct image annotates all of the monitors and the desk where they are located. In the prediction results, although there is only one, the monitor is surrounded by a blue frame and the AnsFields prediction is close to the correct image.

Finally, we show an example when images used for model training are annotated with bounding boxes and drawn with the top point highlighting. The locations of ``ottomans'' and ``coaches'' need to be properly predicted from within the scene to identify where they can be seen well. 
Some parts of bounding box prediction were successful and the result of AnsFields are predicted ``ottomans'' and ``coaches'' were well visible.

Thus, we have successfully used the diffusion model to understand the character of the objects and their location in the scene needed to answer a question and to predict the appropriate location to answer the question.

\section{Conclusion}
\label{sec:conclusion}
This paper introduced Answerability Fields as a novel approach to predicting answerability within complex indoor environments. Leveraging data from the ScanQA dataset, we constructed a comprehensive AnswerabilityFields dataset encompassing diverse scenes and questions from ScanNet. Utilizing a diffusion model, we successfully inferred and evaluated these Answerability Fields, shedding light on objects' importance and locations in answering questions within a scene.

Our results highlight the efficacy of our approach in guiding scene navigation tasks by utilizing the gradient of Answerability Fields. By demonstrating the potential application of this method, we pave the way for enhanced interactions between intelligent agents and their environments. Moving forward, we envision further advancements and applications of Answerability Fields in various domains, ultimately contributing to the advancement of embodied artificial intelligence.

\bibliographystyle{plain}
\bibliography{main}

\end{document}